\documentclass[conference]{IEEEtran}
\IEEEoverridecommandlockouts
\usepackage{cite}
\usepackage{amsmath,amssymb,amsfonts}
\usepackage{algorithmic}
\usepackage{graphicx}
\usepackage{textcomp}
\usepackage{xcolor}
\usepackage{multirow}
\usepackage{booktabs}
\usepackage{changes}
\usepackage{soul}
\usepackage{xcolor}
\def\BibTeX{{\rm B\kern-.05em{\sc i\kern-.025em b}\kern-.08em
    T\kern-.1667em\lower.7ex\hbox{E}\kern-.125emX}}
\usepackage{url}
\usepackage{algorithm}
\usepackage{enumitem}
\usepackage{balance}
\usepackage{times} % or
\usepackage{mathptmx}

\begin{document}

\title{Haphazard Inputs as Images in Online Learning
\thanks{This work was supported by the Research Council of Norway Project (nanoAI, Project ID: 325741), H2020 Project (OrganVision, Project ID: 964800), ERC PoC Spermotile (Project ID: 101215323) and EIC Transition Spermotile (Project ID: 101123485).}
}

\author{\IEEEauthorblockN{Rohit Agarwal}
\IEEEauthorblockA{\textit{Department of Computer Science} \\
\textit{UiT The Arctic University of Norway}\\
Tromso, Norway \\
agarwal.102497@gmail.com}
\and
\IEEEauthorblockN{Aryan Dessai}
\IEEEauthorblockA{\textit{Department of Mathematics and Computing} \\
\textit{IIT (ISM) Dhanbad}\\
Dhanbad, India \\
dessaiaryan4419@gmail.com}
\and
\IEEEauthorblockN{Arif	Sekh}
\IEEEauthorblockA{\textit{Department of Computer Science} \\
\textit{UiT The Arctic University of Norway}\\
Tromso, Norway \\
arif.a.sekh@uit.no}
\and
\IEEEauthorblockN{Krishna	Agarwal}
\IEEEauthorblockA{\textit{Department of Physics and Technology} \\
\textit{UiT The Arctic University of Norway}\\
Tromso, Norway \\
krishna.agarwal@uit.no}
\and
\IEEEauthorblockN{Alexander	Horsch}
\IEEEauthorblockA{\textit{Department of Computer Science} \\
\textit{UiT The Arctic University of Norway}\\
Tromso, Norway \\
alexander.horsch@uit.no}
\and
\IEEEauthorblockN{Dilip	K. Prasad}
\IEEEauthorblockA{\textit{Department of Computer Science} \\
\textit{UiT The Arctic University of Norway}\\
Tromso, Norway \\
dilipprasad@gmail.com}
}

\maketitle

\begin{abstract}
The field of varying feature space in online learning settings, also known as haphazard inputs, is very prominent nowadays due to its applicability in various fields. However, the current solutions to haphazard inputs are model-dependent and cannot benefit from the existing advanced deep-learning methods, which necessitate inputs of fixed dimensions. Therefore, we propose to transform the varying feature space in an online learning setting to a fixed-dimension image representation on the fly. This simple yet novel approach is model-agnostic, allowing any vision-based models to be applicable for haphazard inputs, as demonstrated using ResNet and ViT. The image representation handles the inconsistent input data seamlessly, making our proposed approach scalable and robust. We show the efficacy of our method on four publicly available datasets. The code is available at \url{https://github.com/Rohit102497/HaphazardInputsAsImages}.

\end{abstract}


\section{Introduction}
Online learning involves updating the model continuously as new data arrives in a data stream. The model learns from each new data point rather than being trained on the whole dataset at once \cite{OnlineLearning}. It is beneficial in scenarios where data is continuously generated, such as real-time recommendation systems, fraud detection, autonomous vehicles, modeling sensor networks, etc. \cite{OLSurvey}. The primary challenge of such a system is that the model might become unstable if the new data is noisy or not representative of the overall data distribution \cite{ConceptDrift}.

On the other hand, haphazard inputs, a complex case of online learning, refer to data that arrives inconsistently, with varying dimensions, missing features, or unexpected new features \cite{auxdrop}. For example, in Internet of Things (IoT) systems, sensors often send data that can be inconsistent due to network issues or sensor faults. Similarly, medical data from various sources (e.g., wearable devices and electronic health records) can be incomplete, irregular, and evolving over time. Handling haphazard inputs is challenging due to inconsistent data dimensions, missing or noisy data, the sudden appearance of new features, or the obsoleteness of old features. These challenges require robust models and sophisticated techniques to ensure accurate and reliable performance \cite{UtilitarianSurvey}.

We envision an interesting future application of haphazard inputs in space and maritime exploration, where numerous unforeseen variables and rapid changes prevail, requiring on-the-fly adaptions. The dynamic scalability facilitated by models on haphazard inputs could revolutionize operational protocols, enhancing efficiency and safety while conserving resources.

\begin{figure}[t]
    \begin{center}
    \includegraphics[width=0.48\textwidth, trim={1.0cm 1.0cm 1.8cm 0.0cm}, clip]{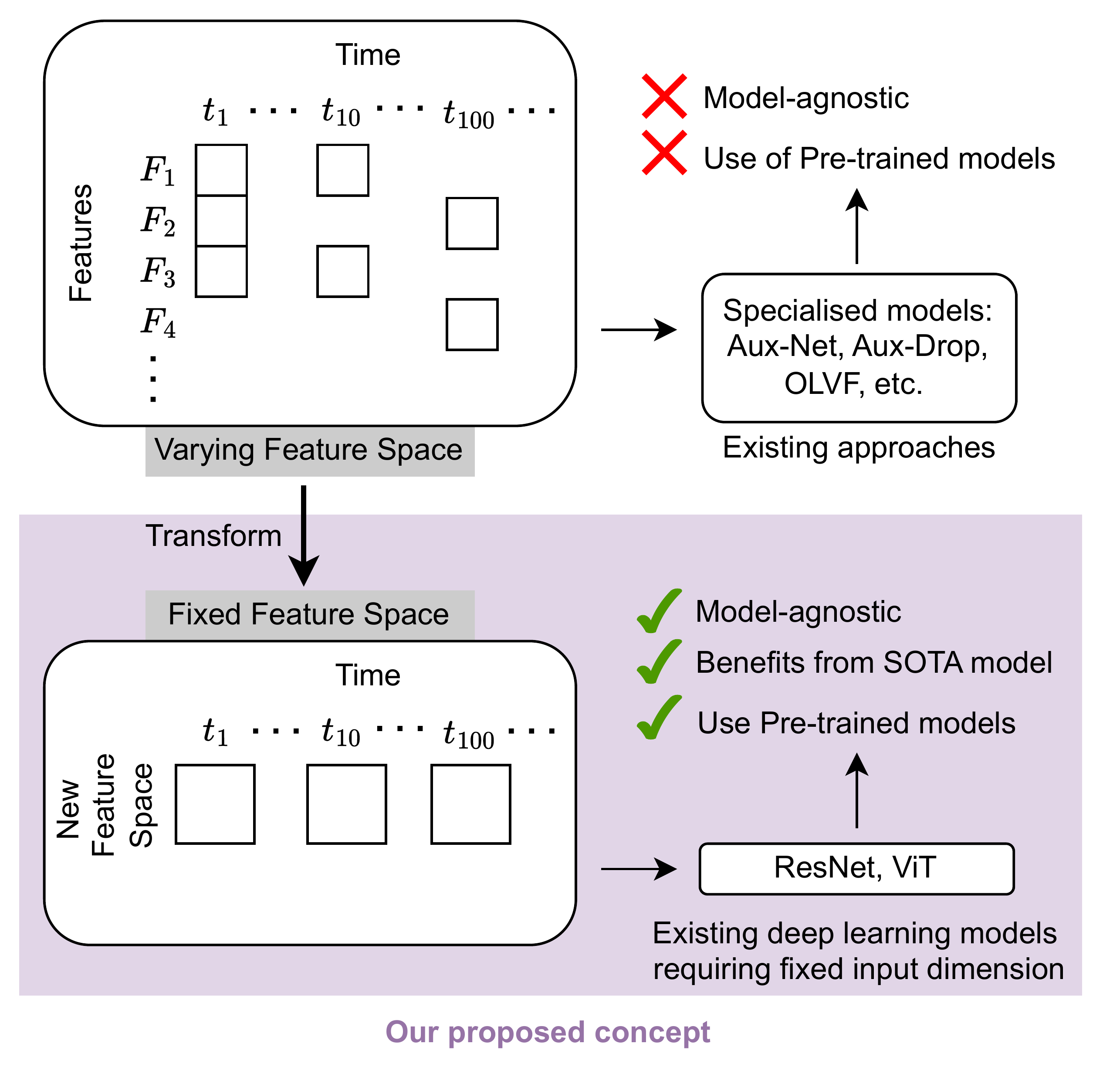} 
    \end{center}
    \caption{The advantages of the proposed solution over existing models in the haphazard inputs field (or varying feature space in online learning).}
    \label{fig:teaser}
\end{figure}

Recent progress in haphazard inputs, such as Aux-Net \cite{auxnet} and Aux-Drop \cite{auxdrop}, provide a promising baseline. However, they have their limitations. Aux-Net has a high time and space complexity, making it inefficient and not scalable for higher dimensional data due to architectural limitations. It struggles to quickly adapt to new or missing features. Aux-Drop relies heavily on dropout regularization to handle haphazard inputs. While this helps prevent the co-adaptation of features, it may not always be sufficient to manage highly chaotic data inputs. Most importantly, the current solutions to haphazard inputs are model-dependent and cannot benefit from the existing advanced deep-learning methods (see Fig. \ref{fig:teaser}). 

Inspired by the above challenges, we propose a novel model-agnostic approach to handle haphazard inputs. The key idea is to transform the variable feature space of haphazard inputs to a fixed dimensional feature space, facilitating the utilization of the existing deep learning models. Specifically, we convert the one-dimensional (1D) varying feature space into a fixed-dimension bar plot image (2D). A computer vision (CV) based neural network is then employed to perform the classification in an online setting. We show the efficacy of our approach on four publicly available benchmark datasets.

The contribution of our method lies in addressing several challenges of haphazard inputs: 
\begin{enumerate}[label=(\alph*)]
    \item \textbf{Enhanced Scalability:} The image representation of a varying number of features allows our approach to scale effectively, accommodating any number of features.
    \item \textbf{Model Agnostic:} The fixed space transformation of haphazard inputs facilitates the use of any CV-based model, making our approach a model-agnostic concept. 
    \item \textbf{Robustness to Unobserved Data:} The bar plot representation can easily manage missing features, sudden features, obsolete features, missing data, and an unknown number of total features, ensuring reliable performance.
\end{enumerate}

\noindent Our approach is the first novel method to use CV-based models in modeling haphazard inputs. This paradigm shift leverages the strengths of image-based classifiers to handle the irregularities and inconsistencies inherent in haphazard inputs, providing a more efficient and scalable solution.

\section{Related Works}

Our article aligns with three distinct research trajectories: (1) online learning, which raises the challenges of haphazard inputs, (2) the problem of haphazard inputs, which we tackle in this paper, and (3) the transformation of raw data to images, which forms the basis of our space transformation from variable space to fixed space.

\subsection{Online Learning}
Online learning has been handled using many classical methods like decision trees \cite{DecisionTrees}, support vector machines \cite{svm}, Bayesian theory \cite{Bayesian}, and fuzzy logic \cite{fuzzy}. Incremental learning approaches have also seen prominence in online learning \cite{IncrementalMF}. However, all these methods work on the assumption of the fixed input feature space. Some recent research fields like feature evolvable streams \cite{FeatureEvolvable}, incremental and decremental features \cite{IncrementalDecremental}, trapezoidal data streams \cite{Trapezoidal}, and unpredictable feature evolution \cite{UnpredictableFeature} partially alleviate this assumption \cite{HaphazardInputSurvey}. Still, these methods assume either batch learning or some form of structure in their data, making them inapplicable to the field of haphazard inputs.

\subsection{Haphazard Inputs} Haphazard inputs were first addressed by Katakis et al. \cite{NB3} in 2005, who proposed naive Bayes (NB3) to incorporate dynamic features. Feature Adaptive Ensemble (FAE) \cite{FAE} expands NB3 by incorporating an ensemble of naive Bayes classifiers. In recent times, methods like Online Learning from Capricious Data Streams (OCDS) \cite{OCDS} and Online Learning in Variable Feature Spaces with Mixed Data (OVFM) \cite{OVFM} proposed reconstruction methods to determine the values of unobserved features from observed features. Online Learning from Varying
Features (OLVF) \cite{OLVF} explored utilizing empirical risk minimization to project haphazard inputs into a shared subspace. Online Learning for Data Streams with Incomplete Features and Labels (OLIFL) \cite{OLIFL} learns from each feature by maintaining an informative matrix. Dynamic Forest (DynFo) \cite{DynFo} and Online Random Feature Forests for Feature space Variabilities (ORF\textsuperscript{3}V) \cite{ORF3V} showed that simple solutions like decision stumps also give competitive performance. 

Since the above methods are classical models, they perform well in smaller datasets. However, for large datasets, it is important to develop deep learning models. With this motive, Aux-Net \cite{auxnet} and Aux-Drop \cite{auxdrop} proposed deep learning-based methods. However, these are model-dependent solutions, unable to exploit the power of existing deep neural architectures.

\subsection{Information-to-Image Based Learning}
Transforming non-image data, such as tabular, audio, and time-series data, into images for classification tasks has gained significant traction. This trend is primarily driven by the nature of image-based convolutions, which preserve spatial relationships and allow for a comprehensive representation of information. By converting data into images, researchers can leverage data-agnostic and model-agnostic approaches and utilize the power of pre-trained models.

In this context, Sharma et al. \cite{DeepInsight} introduced a method to convert tabular data into t-SNE plots, which were then used for classification tasks. This method, applied to sRNA datasets, achieved state-of-the-art accuracy. Similarly, Zhu et al. \cite{IGTD} proposed representing each data point as a pixel in an image, while Damri et al. \cite{FCViz} used a feature correlation structure. Li et al. \cite{ViTST} suggested converting time-series data into line plots for classification purposes. Velarde et al. \cite{Audio} employed a symbolic image-based representation of music for classification tasks, reporting superior performance. Ryan et al. \cite{AnomalyDetection1} handled anomaly localization by transforming sequential data to an image, using a specialized filter that can produce flexible shape forms and detect multiple types of outliers simultaneously. Recently, Kang et al. \cite{AnomalyDetection2}  proposed a time-series to image-transformed adversarial autoencoder to effectively capture the local features of adjacent time points.

However, it is important to note that all the above methods are offline learning benefiting from the batches of data and multi-pass training. Therefore, these methods are not applicable to handle the challenges of variable feature space posed by haphazard input in an online learning setting.

\section{Problem Statement} 
\label{sec:problem_statement}

Haphazard inputs are streaming data with varying numbers of available features at each instance and an unknown number of total features. The problem formulation is given by \(f^{t-1}: X^t \rightarrow Y^t\), where \(X^t \in \mathbb{R}^{d^t}\) is the haphazard input received at time instance \(t\), and \(Y^t\) represents the ground truth at time \(t\). Here, \(d^t\) represents the variable number of features available at time \(t\). The numerical value of an available feature \(j\) at time \(t\) is represented by \(x^t_j\). The prediction problem in haphazard inputs can be both regression and classification. In this work, aligned with previous works like \cite{OLVF} and \cite{auxdrop}, we focus on binary classification. Therefore, \(Y^t \in \{0, 1\}\). Note that the work here can be easily extended to multi-class classification and regression problems. The \(f^t\) denotes the model \(f\) learned till time \(t\). \(f^0\) is the initialized model. The model learns in an \textbf{online learning setting}. The received input \(x^t\) at time \(t\) is processed by \(f^{t-1}\) to give a prediction \(\hat{Y}^t\). The model calculates the loss based on the ground truth \(Y^t\) and the predicted output \(\hat{Y}^t\). The loss is given by \(L^t = G(Y^t, \hat{Y}^t)\), where \(G\) is the loss function. The model \(f\) updates its parameters based on \(L^t\) to give an updated model \(f^t\).

Haphazard inputs are characterized by six properties \cite{auxdrop} -- \textbf{(a) Streaming data}: The data arrives one instance after another. Since the data is huge and cannot be stored, online learning needs to be employed to model streaming data. \textbf{(b) Missing data}: Some of the features may not be observed because of various reasons like faulty sensors, network failure, etc. These features contribute to missing data. \textbf{(c) Missing features}: Some of the features may not be available from the onset. But it is known that they will be available in future instances. These features are known as missing features. \textbf{(d) Sudden features}: Some of the features arrive in future instances. However, their presence is unknown from the onset. \textbf{(e) Obsolete features}: The features become unavailable for all future instances after a certain time period. These features are known as obsolete features. Note that the cessation of obsolete features may not be known. \textbf{(f) Unknown number of total features}: Due to the missing data, missing features, sudden features, and obsolete features characteristics of haphazard inputs, the total number of features is unknown. In this work, we term all the features available at any instance as ``\textit{observed features}". Other features that are not seen at a time instance owing to missing data, missing features, sudden features, and obsolete features are termed ``\textit{unobserved features}".

The advancements in deep learning have been significant. However, the haphazard inputs cannot be modeled by the existing deep-learning architectures because they require the input dimension to be fixed. Hence, the current landscape of haphazard inputs involves creating new and scalable architectures. But, in this work, we take a step back and propose a solution that can benefit from the existing deep learning architectures. The proposed solution is discussed next.

\begin{figure*}[t]
    \begin{center}
    \includegraphics[width=\textwidth, trim={0.0cm 0.0cm 0.0cm 0.0cm}, clip]{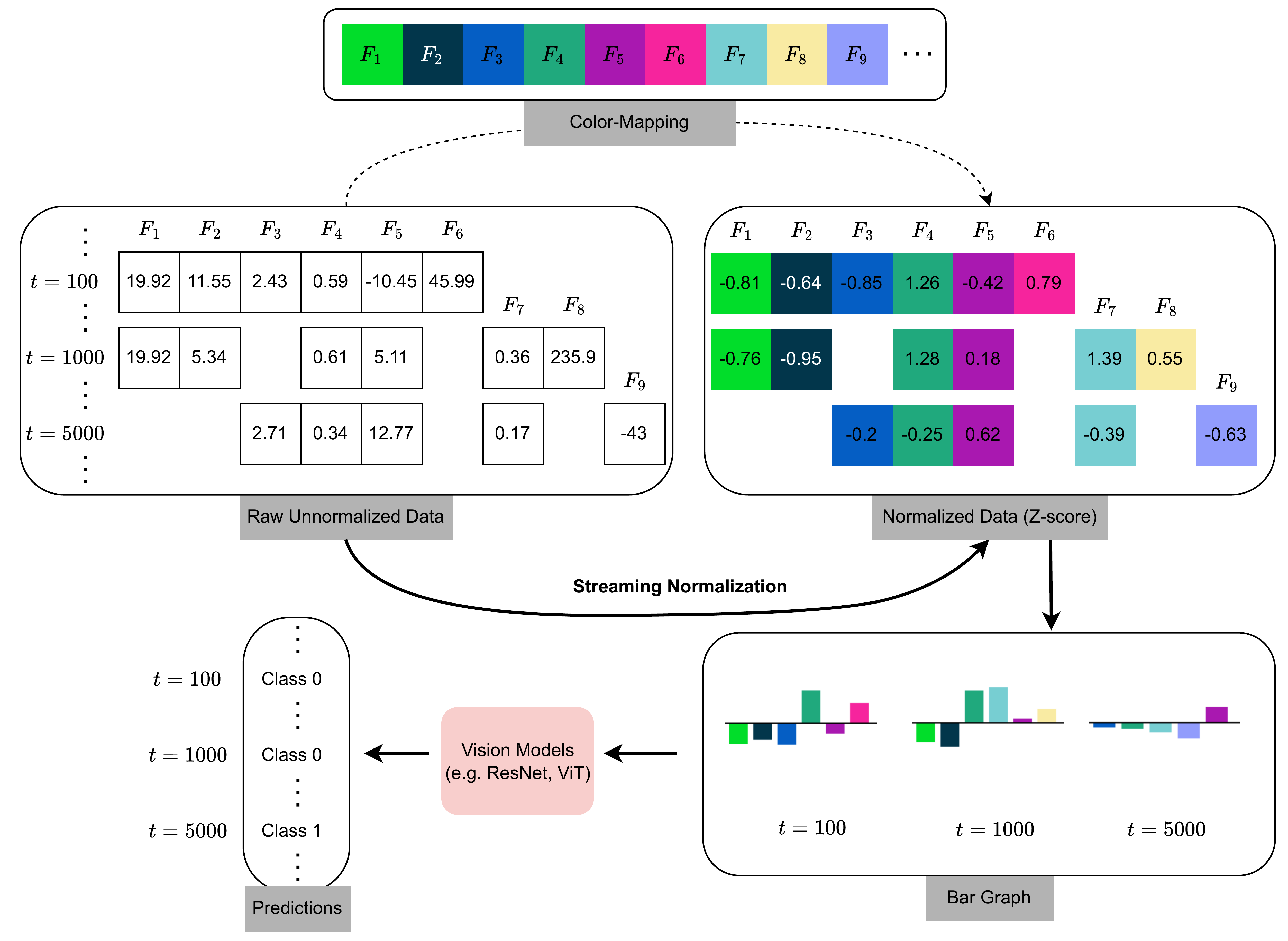} 
    \end{center}
    \caption{HI\textsuperscript{2} concept: Initially, we receive the streaming data as shown by a snapshot of magic04 data (see Table \ref{tab:data_description}) in the first box. The values in this figure are rounded to 2 decimal digits for ease of visualization. Next, for each new feature, a unique color is generated in the color-mapping storage. The colors corresponding to the observed features are selected for image creation. Subsequently, the raw data is normalized in a streaming manner. The normalized features are then converted into a bar graph of 224\(\times\)224 dimensions with corresponding features color. Finally, an image classifier is used to give a prediction.}
    \label{fig:method}
\end{figure*}
\section{Proposed Solution}

The deep learning architectures assume the input to be of fixed dimension, i.e., \(I^t \in \mathbb{R}^d \) \(\forall \) \(t\). However, the haphazard inputs have variable dimensions, i.e., \(X^t \in \mathbb{R}^{d^t}\). Therefore, our aim to utilize deep learning methods for haphazard inputs necessitates the transformation of variable input space to fixed dimension representation, i.e., \(X^t \rightarrow I^t\). 

The ease of representing numerical values in the form of graphs and the comprehensive representation of information motivated us to consider image representation for the transformation. The image representation also allows to accommodate any number of features making the proposed solution scalable. Furthermore, it allows the applicability of CV-based models, benefitting from their pre-trained learning.

\textit{Thus, we propose a novel model-agnostic concept, termed \underline{H}aphazard \underline{I}nputs as \underline{I}mages (HI\textsuperscript{2}), pictorially representing the haphazard inputs in the form of images, allowing to utilize CV-based models to handle haphazard inputs.} In other words, HI\textsuperscript{2} converts the raw numerical information from the data entries to images that represent data in terms of comparison charts (e.g., bar charts), highlighting the relative difference in magnitudes of different feature values. Vision models are then trained to infer from these visual features. Therefore, HI\textsuperscript{2} converts the problem of haphazard inputs into a vision-based task in an online learning setting, overcoming the handicap of deep learning models posed by varying input feature space. The concept of HI\textsuperscript{2} is depicted in Fig. \ref{fig:method} and discussed next.

\subsection{Image Transformation} 
\label{sec:ImageTransformation}

We represent haphazard inputs in the form of bar graph images. Each feature is uniquely identified by a color. Each color is uniquely generated by different permutations of RGB color values and stored for future reference. The RGB values are randomly generated with a fixed seed so that the feature-color mapping is consistent throughout the model's training and testing in an online learning setting. Whenever a new feature is observed, a unique color is assigned to it, which persists for the entirety of the model.

Instead of the raw numerical values of the features, the normalized feature values are represented as the height of the bars. Since the images represent the relative distinctions in the magnitude of different features, normalization is ideal as it achieves scaling consistency without altering the motive of said representation. We perform Z-score streaming normalization (as discussed in section \ref{sec:zscore}) and clip the value outside the range of [-3, 3] since the values away from 3 standard deviations would be outliers \cite{Outlier}. Therefore, the y-axis in the images ranges from -3 to +3. Instead of Z-score, min-max normalization can also be utilized, as presented in section \ref{sec:minmax}.

Only observed features are plotted at each time. The unobserved features are not considered for representation in the image. This allows the model to accommodate many features effectively in the entirety of the model learning. It is seen in the field of haphazard inputs that the observed features are sparse \cite{UtilitarianSurvey}. Therefore, the number of observed features at each instance is few, although the total number of features (observed + unobserved) can be very high. Since we only represent the observed features in the image, we can accommodate a high number of total features. However, in scenarios where the number of features arriving at a time instance is large, it is handled by constructing a larger image canvas and then reshaping it to 224\(\times\)224 sized images for pre-trained vision models. Note that reshaping may, in some situations, lead to loss of information. However, we note that the observed number of features at each time instance is small enough to not require a large canvas.

The widths of the bars vary across images depending on the number of observed features at each time instance. We include blank spacing between the bars for a separating boundary between two consecutive bars. This ensures an easy distinction between the represented features. Thus the bar width (\(B_w^t\)) is dynamic and depends upon the number of observed features \(d^t\) at time \(t\).  Naturally, the spacing between bars (\(S^t\)) is also dynamic. We set the spacing as a fixed fraction (\(f_f\)) of  \(B_w^t\). Therefore, at each time instance, \(B_w^t\) and \(S^t\) are given by

\begin{equation}
\label{eq:barwidth}
    \begin{split}
        B_w^t &= int(224 /(d^t + (d^t + 1)f_f)) \\
        S^t &= int(B_w^t * f_f)   
    \end{split}
\end{equation}

This transforms the varying dimension of haphazard input \(X^t \in \mathbb{R}^{d^t}\) to an image \(I^t \in \mathbb{R}^{3\times224\times224}\) of fixed dimension. A few examples of the transformer bar graphs are in Fig. \ref{fig:method}.

\subsection{Streaming Normalization}
\label{sec:normalization}

The feature values can vary significantly in magnitude, necessitating normalization for consistent representation. However, haphazard inputs operate in an online learning setting. Therefore, batch or entire data normalization is not feasible. We normalize based on the data seen so far by determining the running statistics \cite{running_stats} of each feature. Out of the many normalization techniques in the literature, we experiment with two popular ones: Z-score and Min-Max Normalization.

\begin{algorithm}[tb]
   \caption{Haphazard Inputs as Images}
   \label{alg:method}
\begin{algorithmic}
    
   \STATE {\bfseries Input:} CV model \(C\) with learnable parameters \(\theta\) at time \(t-1\), i.e. \(C^{t-1}_{\theta^{t-1}}\), color-mapping storage \(\{<feature,color>\}\), and haphazard input \(X^t\)\\
   \STATE {\bfseries Output:}  \(C^{t}_{\theta^{t}}\), \(\{<feature,color>\}\)
   \STATE {\bfseries Procedure:} Learning at time \(t\)
    \STATE{\(feature_{new} = \{feat_j\} \hspace{1em} \forall \hspace{0.3em} feat_j \in \{X^t  - feature\)\}} 
    \STATE{ \(color_{new} = \{random(j)\} \hspace{1em} \forall \hspace{0.3em} feat_j \in feature_{new}\)}
    \STATE{ \(\{<feature, color>\}= \{<feature,color>\} \hspace{0.3em} \cup\)}
    \STATE{\hspace{11.0em} \(\{<feature_{new}, color_{new}>\}\)} 
    \STATE \(X^{t}_{normalize}\) = Normalize \(X^t\) using eq. \ref{eq:Zscore} or \ref{eq:minmax}.
    \STATE {\(\{B^t_w,S^t\}\) = Calculate using eq. \ref{eq:barwidth}.} 
    \STATE \(I^t\) = Generate using \(B^t_w\), \(S^t\), and \(X^{t}_{normalized}\)
    \STATE \(\hat{Y}^t\) = Prediction using eq. \ref{eq:prediction} 
    \STATE \(L^t\) = Calculate loss using eq. \ref{eq:loss}.
    \STATE \(C^{t}_{\theta^{t}}\) = Update model \(C^{t-1}_{\theta^{t-1}}\) using eq. \ref{eq:gradientdescent}         
    \STATE{{\bfseries Return:} \(C^{t}_{\theta^{t}}\), \(\{<feature,color>\}\)}
\end{algorithmic}
\end{algorithm}

\paragraph{Z-Score Normalization} 
\label{sec:zscore}
We utilize the Knuth et al. \cite{running_stats} method of determining the running statistics because of its numerical stability. Let us denote the running mean and running standard deviation of feature \(j\) till time \(t\) by \(\mu^t_j\) and \(\sigma^t_j\), where 
\begin{equation}
    \begin{split}
        \mu^t_j &= \mu^{t_-}_j + \frac{x_{j}^{t} - \mu_{j}^{t_-}}{k_j^t}, \text{ and}\\
        ({\sigma_{j}^{t}})^2 &= \frac{v^t_j}{k_j^t-1}.
    \end{split}
\end{equation}
Here, \(t_-\) represents the time instance prior to \(t\), when the feature $j$ was observed. The \(k_j^t\) represents the number of times feature \(j\) was available till time \(t\). The \(v^t_j = v^{t_-}_j + (x^t_j - \mu^{t_-}_j)(x^t_j - \mu^t_j)\). Finally, the Z-score of \(j^\text{th}\) feature at time \(t\) is given by 
\begin{equation}
    \label{eq:Zscore}
    \frac{x_{j}^{t}-\mu_{j}^{t}}{\sigma_{j}^{t}}.
\end{equation}

\paragraph{Min-Max Normalization}
\label{sec:minmax}
The feature values are normalized to [0, 1] intervals using the minimum (\(Pn_{j}^{t}\)) and maximum value (\(Px_{j}^{t}\)) of each feature from the past data as
\begin{equation}
    \label{eq:minmax}
    \frac{x_{j}^{t} - Pn_{j}^{t}}{Px_{j}^{t} - Pn_{j}^{t}}. 
\end{equation}
The $Px_{j}^{t}$ and $Pn_{j}^{t}$ are updated in an online manner as and when the new value of feature $j$ becomes available at time \(t\). Note that, in the case of Min-Max normalization, the y-axis of the transformed images will range from 0 to 1 (inclusive).

\subsection{Vision models}
The transformer image \(I^t\) has two main properties, i.e., the color of the bars representing different features and the height of the bar denoting feature values. Therefore, we utilize vision models as they can effectively learn from the color representation and shape, also shown in previous articles \cite{vision}.

Let us represent a CV-based classifier by \(C_\theta\), where \(\theta\) denotes the learnable parameters. Then, the vision task is to learn \(C^{t-1}_{\theta^{t-1}}: I^t \rightarrow Y^t\), where \(C^{t-1}_{\theta^{t-1}}\) denotes the classifier learned till time \(t-1\). The output of the classifier is given by
\begin{equation}
    \label{eq:prediction}
    \hat{Y}^t = C^{t-1}_{\theta^{t-1}}(I^t).
\end{equation}
The loss function (represented by \(G\)) determines the loss of the classifier at time \(t\) (\(L^t\)) as
\begin{equation}
    \label{eq:loss}
    L^t = G(Y^t, \hat{Y}^t).
\end{equation}
The classifier updates its parameters in an online gradient descent \cite{OnlineGradientDescent} manner as
\begin{equation}
    \label{eq:gradientdescent}
    \theta^t = \theta^{t-1} - \eta\frac{\delta{L^t}}{\delta{\theta^{t-1}}}.
\end{equation}
The algorithm of the HI\textsuperscript{2} approach is detailed in Algorithm \ref{alg:method}.

\section{Experiments}

\begin{table}[t]
\caption{Datasets Description: No. of \# and Imbalance Ratio denotes the number of \# and positive labels in the whole dataset.}
\begin{center}
\begin{tabular}{|c|c|c|c|}
\hline
\textbf{Dataset} & \textbf{No. of Features} & \textbf{No. of Instances} & \textbf{Imbalance Ratio} \\

\hline
magic04 & 10 & 19020 & 64.84\% \\
a8a & 123 & 32561 & 75.92\% \\
SUSY & 8 & 1 Million &  45.79\%  \\
HIGGS & 21 & 1 Million  & 52.97\% \\
\hline
\end{tabular}
\label{tab:data_description}
\end{center}
\end{table}

\subsection{Datasets} We evaluate our method on four datasets, namely, magic04 \cite{magic04}, a8a \cite{a8a}, SUSY \cite{susyhiggs}, and HIGGS \cite{susyhiggs}. The description of these datasets is provided in Table \ref{tab:data_description}. These four datasets encapsulate varying numbers of features and instances, where the number of features ranges from 8 (in SUSY) to 123 (in a8a), and the number of instances ranges from 19020 (in magic04) to 1 Million (in SUSY and HIGGS). Therefore, these datasets provide enough variability to assess the efficacy of HI\textsuperscript{2}. The datasets also have both balanced and imbalanced data, with a8a being the most imbalanced with 75.92\% positive labels and HIGGS being the balanced dataset with 52.97\% positive labels. We consider varying levels of unavailable datasets to better evaluate our method on different data availability. Specifically, we test on 25\%, 50\%, and 75\% unavailable data.

\paragraph{Creating haphazard inputs} Each of the four datasets is transformed into three different datasets with varying levels of data availability. Specifically, we drop \(100\times(1-p)\%\) of total features at each time instance, as done in previous baseline papers \cite{OLVF, auxdrop}. Therefore, \(100\times{p}\%\) of features are simulated as observed independently of other features following a uniform distribution at each time instance. We create three sets of data, each at \(p = 0.25\), \(0.50\), and \(0.75\). 

\begin{table*}[htbp]
\caption{Performance comparison of HI\textsuperscript{2} and baseline models. The deterministic models like NB3, FAE, OLVF, and OLIFL were run only once, whereas non-deterministic models were executed five times, and mean \(\pm\) standard deviation was reported. Some of the non-deterministic models could be run only once due to their significant time requirements and are denoted by \textsuperscript{$\ddagger$} symbol.}
\begin{center}
\resizebox{\linewidth}{!}{%
\begin{tabular}{lcccccccccccc}
\toprule
& & \multicolumn{8}{c}{\textbf{Classical Models}} & \multicolumn{3}{c}{\textbf{Deep-Learning Models}} \\
\cmidrule(r){3-10}
\cmidrule(l){11-13} 
Dataset & $p$ & NB3 & FAE & OLVF & OLIFL & OCDS & OVFM & DynFo & ORF\textsuperscript{3}V & Aux-Net & Aux-Drop & \textbf{HI\textsuperscript{2}} \\
\midrule

\multicolumn{13}{c}{\textbf{\underline{Balanced Accuracy}}} \vspace{0.5em} \\

\multirow{3}{*}{magic04} & 
  0.25 & 50.01 & 50.01 & 53.18 & 53.06 & 51.89$\pm$0.10 & 51.94$\pm$0.00 & 52.75$\pm$0.30	& 47.94$\pm$0.22 & 50.09$\pm$0.07 & 56.04$\pm$0.53 & \textbf{59.89$\pm$0.50} \\
  &  0.50 & 50.02 & 50.00 & 54.60 & 57.28 & 53.40$\pm$0.45	& 54.13$\pm$0.08 &	55.12$\pm$0.06	& 48.56$\pm$0.11 & 50.09$\pm$0.03 & 59.29$\pm$0.48 & \textbf{67.09$\pm$0.53} \\ 
  &  0.75 & 49.99 & 50.00	& 56.19	& 60.75 & 53.76$\pm$1.07 & 58.79$\pm$0.04	& 56.75$\pm$0.02 & 49.32$\pm$0.04 & 50.05$\pm$0.07 & 63.18$\pm$0.61 & \textbf{74.28$\pm$0.11} \\
\midrule

\multirow{3}{*}{a8a} & 
   0.25& 50.01 & 50.00	& \textbf{60.67} & 53.57 & 54.75$\pm$0.87 & 58.66$\pm$0.00 & 50.01$\pm$0.03 & 49.99$\pm$0.00 & 50.00$\pm$0.00 & 50.00$\pm$0.01 & 56.11$\pm$0.43 \\
  &  0.50 & 50.01 & 50.00 &  \textbf{66.46}  & 54.63 & 64.04$\pm$1.01 & 66.02$\pm$0.00 & 50.11$\pm$0.01 & 50.01$\pm$0.00 & 50.00$\pm$0.00 & 55.33$\pm$1.99 & 65.20$\pm$0.22 \\
  &  0.75& 50.01 & 50.00	 & 70.60  & 56.56 & 68.81$\pm$1.10 & \textbf{70.95$\pm$0.00}	& 50.13$\pm$0.01 & 49.99$\pm$0.00	& 50.00$\pm$0.00 & 62.87$\pm$0.93 & 70.55$\pm$0.31 \\
\midrule

\multirow{3}{*}{SUSY} & 
   0.25 & 50.00 &   49.90    & 51.12  & 51.23	 & 52.11$\pm$0.19 & 58.00$\pm$0.00 & 54.69$\pm$0.01 & 49.37$\pm$0.01	& 50.53$\pm$1.17 & 61.98$\pm$0.10 & \textbf{62.55$\pm$0.04}\\
  &  0.50 & 50.00	& 50.01 & 53.21	& 53.59 & 54.03$\pm$0.28 & 62.85$\pm$0.00 & 58.27$\pm$0.00 & 48.33$\pm$0.02 & 57.89$\pm$7.19 & 68.79$\pm$0.14 & \textbf{69.31$\pm$0.03} \\
  &  0.75 & 50.00 & 50.12	& 55.98	& 56.39 & 54.84$\pm$0.48 & 68.51$\pm$0.00	& 60.94$\pm$0.01 & 47.53$\pm$0.03 & 53.67$\pm$8.13 & 73.55$\pm$0.11 & \textbf{74.17$\pm$0.02} \\
\midrule

\multirow{3}{*}{HIGGS} & 
   0.25 & 50.00 &   50.16 & 50.57 & 50.56 & 49.97$\pm$0.07	& 50.61$\pm$0.01 & 50.18\textsuperscript{$\ddagger$} & 49.86$\pm$0.03 & 49.99$\pm$0.00 & \textbf{51.17$\pm$0.05} & 50.39$\pm$0.20\\
  &  0.50 & 50.00 & 50.01 & 51.21	& 51.50 & 50.06$\pm$0.06 & 51.40$\pm$0.01 & 50.21\textsuperscript{$\ddagger$} & 49.82$\pm$0.02 & 49.99$\pm$0.01	& 53.09$\pm$0.05 & \textbf{53.89$\pm$0.25} \\
  &  0.75& 50.00	& 50.55	& 51.98	& 52.48 & 49.97$\pm$0.05 & 52.66$\pm$0.00	& 50.16\textsuperscript{$\ddagger$} & 49.75$\pm$0.03 & 49.98\textsuperscript{$\ddagger$} & 55.55$\pm$0.11 &	\textbf{58.12$\pm$0.12} \\
\midrule

\multicolumn{13}{c}{\textbf{\underline{AUPRC}}} \vspace{0.5em} \\
\multirow{3}{*}{magic04} & 
  0.25 & 64.73 & 62.53 & 59.34 & 68.63 & 62.96$\pm$0.22	& 70.14$\pm$0.00 & 65.56$\pm$0.11	& 63.04$\pm$0.05 & 64.80$\pm$0.19	& 68.46$\pm$0.59 & \textbf{76.65$\pm$0.43} \\
  &  0.50 & 65.21 &	61.36 & 60.29 & 67.54 & 66.40$\pm$1.20	& 72.72$\pm$0.06 &	67.65$\pm$0.02	& 62.24$\pm$0.09 & 64.61$\pm$0.22	& 71.25$\pm$0.31 & \textbf{84.20$\pm$0.34} \\ 
  &  0.75 & 64.15 & 59.67	& 61.4	& 64.53 &  67.60$\pm$1.39 & 77.80$\pm$0.05	& 67.60$\pm$0.01 & 59.68$\pm$0.14 & 64.88$\pm$0.33	& 75.93$\pm$0.43 &  \textbf{89.46$\pm$0.29} \\
\midrule

\multirow{3}{*}{a8a} & 
   0.25& 77.2 &	75.82 & 85.07  & 66.67 & 82.31$\pm$0.88	& \textbf{91.36$\pm$0.00} & 76.35$\pm$0.06	& 74.49$\pm$0.03 & 77.78$\pm$0.23	& 81.27$\pm$2.08 & 89.36$\pm$0.12 \\
  &  0.50 & 76.55 &	75.84 & 88.6  & 60.82 & 89.40$\pm$0.51	& \textbf{94.12$\pm$0.00} & 76.88$\pm$0.06	& 74.27$\pm$0.02 & 79.40$\pm$0.26	& 89.28$\pm$0.93 &  93.29$\pm$0.04\\
  &  0.75 & 76.16 & 75.84	& 90	 & 55.82 & 91.64$\pm$0.31 & \textbf{95.35$\pm$0.00}	& 77.11$\pm$0.06 & 74.07$\pm$0.03	& 81.26$\pm$0.30 & 92.35$\pm$0.43	&  94.82$\pm$0.08 \\
\midrule

\multirow{3}{*}{SUSY} & 
   0.25 & 45.68 &	46.06   & 57.31	 & 54.51 & 53.64$\pm$0.94 & 59.45$\pm$0.00 & 45.94$\pm$0.00 & 45.23$\pm$0.01	& 46.93$\pm$2.26 & 67.50$\pm$0.09 & \textbf{68.00$\pm$0.02} \\
  &  0.50 & 45.79 &	45.51 & 62.65 & 53.28 & 54.01$\pm$0.65	& 66.74$\pm$0.00 & 45.88$\pm$0.00	& 44.85$\pm$0.01 & 57.49$\pm$10.17 & 75.44$\pm$0.10 & \textbf{76.49$\pm$0.01} \\
  &  0.75 & 45.75 & 46.43	& 65.74	 & 53.12 & 53.39$\pm$0.40 & 74.04$\pm$0.00	& 45.8$\pm$0.00 & 44.28$\pm$0.02	& 50.87$\pm$10.38	& 80.44$\pm$0.11 & \textbf{81.95$\pm$0.04} \\
\midrule

\multirow{3}{*}{HIGGS} & 
   0.25 & 52.97 & 52.98 & 53.1  & 56.91  & 52.82$\pm$0.06	& 53.80$\pm$0.00 & 52.98\textsuperscript{$\ddagger$} & 53.02$\pm$0.01	& 53.01$\pm$0.02 & \textbf{54.93$\pm$0.16} & 53.32$\pm$0.15 \\
  &  0.50 & 52.9 & 53 & 53.29  & 56.59 & 52.86$\pm$0.05	& 54.86$\pm$0.00 & 52.99\textsuperscript{$\ddagger$} & 52.96$\pm$0.01 & 53.01$\pm$0.02	& 57.81$\pm$0.12 &  \textbf{58.62$\pm$0.34} \\
  &  0.75& 52.94 & 53.48 & 53.66 & 56.83 & 52.80$\pm$0.02 & 56.07$\pm$0.00	& 52.91\textsuperscript{$\ddagger$} & 52.89$\pm$0.00 & 53.04\textsuperscript{$\ddagger$} & 60.78$\pm$0.22 &	\textbf{64.13$\pm$0.13} \\
\midrule

\multicolumn{13}{c}{\textbf{\underline{AUROC}}} \vspace{0.5em} \\
\multirow{3}{*}{magic04} & 
  0.25 & 49.91 & 48.66 & 44.02 & 46.93 & 48.79$\pm$0.26	& 57.22$\pm$0.00 & 50.29$\pm$0.15	& 48.00$\pm$0.09 & 50.13$\pm$0.25	& 57.99$\pm$0.76 & \textbf{66.32$\pm$0.48} \\
  &  0.50 & 50.51 &	47.97 & 45.38 & 42.71 & 55.43$\pm$1.54	& 61.15$\pm$0.06 &	53.24$\pm$0.04	& 47.27$\pm$0.07 & 49.90$\pm$0.34	& 62.39$\pm$0.34 & \textbf{76.61$\pm$0.45} \\ 
  &  0.75 & 49.34 & 46.5	& 46.89	& 39.24 & 55.90$\pm$1.74 & 60.95$\pm$0.02	& 53.79$\pm$0.04 & 44.26$\pm$0.13 & 50.16$\pm$0.24	& 68.48$\pm$0.92 & \textbf{84.20$\pm$0.18} \\
\midrule

\multirow{3}{*}{a8a} & 
   0.25 & 51.98 & 49.7  & 63.25 & 46.27 & 61.04$\pm$1.41	& \textbf{77.70$\pm$0.00} & 50.86$\pm$0.09	& 47.88$\pm$0.05 & 54.45$\pm$0.18	& 55.92$\pm$3.39 & 73.62$\pm$0.27 \\
  &  0.50 & 51 & 49.74 & 69.64  & 45.36 & 73.56$\pm$0.97 & \textbf{83.74$\pm$0.00} & 51.96$\pm$0.08 & 47.28$\pm$0.04 & 58.66$\pm$0.65	& 72.34$\pm$2.56 & 81.88$\pm$0.07 \\
  &  0.75 & 50.88 & 49.74	& 72.43	 & 43.44 & 77.80$\pm$0.57 & \textbf{86.79$\pm$0.00}	& 52.04$\pm$0.08 & 46.94$\pm$0.05	& 62.44$\pm$0.51 & 79.66$\pm$1.14 & 85.57$\pm$0.14\\
\midrule

\multirow{3}{*}{SUSY} & 
   0.25 & 49.88 & 50.41   & 56.26	 & 48.77 & 55.93$\pm$0.32 & 57.52$\pm$0.01 & 50.18$\pm$0.00	& 49.42$\pm$0.01	& 51.03$\pm$1.80 & 69.26$\pm$0.12 & \textbf{69.54$\pm$0.02} \\
  &  0.50 & 50.03 &	50.22 & 61.86 & 46.41 & 57.95$\pm$0.32	& 64.34$\pm$0.00 & 50.04$\pm$0.00 & 49.06$\pm$0.01 & 60.14$\pm$8.30 & 76.48$\pm$0.15 & \textbf{77.04$\pm$0.02} \\
  &  0.75 & 49.96 & 49.51	& 65.15	 & 43.61 & 58.41$\pm$0.32 & 72.28$\pm$0.00	& 49.91$\pm$0.00 & 48.36$\pm$0.02 & 55.41$\pm$10.29	& 81.07$\pm$0.07 & \textbf{81.96$\pm$0.04} \\
\midrule

\multirow{3}{*}{HIGGS} & 
   0.25 & 50.03 & 50.09 & 49.99  & 49.44 & 49.96$\pm$0.07	& 51.13$\pm$0.00 & 49.99\textsuperscript{$\ddagger$} & 50.00$\pm$0.01	& 50.04$\pm$0.02 & \textbf{52.36$\pm$0.17} & 50.58$\pm$0.26 \\
  &  0.50 & 49.97 &	50.12 & 50.29  & 48.50 & 50.04$\pm$0.08	& 52.42$\pm$0.00 & 50.01\textsuperscript{$\ddagger$} & 49.98$\pm$0.01 & 50.05$\pm$0.01 & 55.60$\pm$0.06 & \textbf{56.36$\pm$0.37} \\
  &  0.75& 49.95 & 50.1 & 50.66 & 47.52 & 49.98$\pm$0.02 & 54.01$\pm$0.00	& 49.94\textsuperscript{$\ddagger$} & 49.88$\pm$0.01 & 50.06\textsuperscript{$\ddagger$} & 58.96$\pm$0.18 &	\textbf{64.13$\pm$0.13} \\

\bottomrule
\end{tabular}
}
\label{tab:benchmarking}
\end{center}
\end{table*}

\subsection{Baselines} We consider ten baseline models to compare our method. They are classical models -- NB3 \cite{NB3}, FAE \cite{FAE}, DynFo \cite{DynFo}, ORF\textsuperscript{3}V \cite{ORF3V}, OLVF \cite{OLVF}, OCDS \cite{OCDS}, OVFM \cite{OVFM}, and OLIFL \cite{OLIFL} -- and deep-learning models like Aux-Net \cite{auxnet} and Aux-Drop \cite{auxdrop}. The Aux-Net and Aux-Drop operate on the assumption that some of the features are always available. The recent paper \cite{HaphazardInputSurvey} alleviates this assumption using simple solutions; therefore, for a fair comparison, we adopt these solutions. We could not consider a few other baselines because they lack open-source codes, and we were unable to implement them ourselves due to insufficient details or model complexity.

\subsection{Metrics} 
\label{sec:metrics}
We considered three metrics -- Area Under the Precision-Recall Curve (AUPRC) \cite{AUPRC}, Area Under the Receiver Operating Characteristic Curve (AUROC) \cite{AUROC}, and balanced accuracy \cite{BalancedAccuracy} -- 
because most of the datasets are imbalanced. 

The metrics are calculated after the model processes all the instances in an online learning setting as discussed in section \ref{sec:problem_statement}. The ground truth \(Y^t\) and the prediction \(\hat{Y}^t\) of all the instances are used to determine the above-discussed metrics.

\subsection{Implementation Details}
\label{sec:implementation_details}
Our proposed concept is model-agnostic and can benefit from any pre-trained vision model. In this work, we considered ResNet-34 \cite{ResNet} pre-trained on ImageNet as our CV model, owing to its balance in accuracy and complexity.

Some of the baseline models like NB3, FAE, OLVF, and OLIFL are deterministic, i.e., they produce the same output irrespective of the seed value. Therefore, these models were only executed once and their metrics are reported. The non-deterministic models were run five times, and the mean \(\pm\) standard deviation was reported. The standard deviation shows the statistical significance of the reported results. We fix the learning rate heuristically at 2e-5 with \(f_f\) as 0.3. The Z-score normalization and binary cross-entropy loss are considered. All the models were executed on an NVIDIA DGX A100 machine using PyTorch Framework.

The most popular way to create graphical images is using matplotlib \cite{ViTST}. However, we observed that the process of converting the graph generated through matplotlib API to a tensor is time-consuming and inefficient. Therefore, we directly create tensor values from feature values alleviating the need to create .jpg files. This simple adoption decreased the time requirement by 15 times. Specifically, matplotlib required 1034.82 seconds, whereas our implementation needs 65.23 seconds for the magic04 dataset at \(p = 0.50\).

We follow Agarwal et al. \cite{HaphazardInputSurvey} for the implementation details of all the baseline models except OLIFL \cite{OLIFL}, where the original article is considered. The OVFM requires input buffer storage violating the online learning principles. Therefore, we considered a buffer of two instances for a fair comparison.

\subsection{Results} The HI\textsuperscript{2} consistently outperforms the baseline models across various datasets and metrics, as evident from Table \ref{tab:benchmarking}. For instance, in the magic04 dataset, HI\textsuperscript{2} achieves the highest balanced accuracy, AUPRC, and AUROC values. This trend is observed across other larger datasets like SUSY and HIGGS, where HI\textsuperscript{2} demonstrates superior performance in almost all the scenarios. We note that HI\textsuperscript{2} performs particularly well at higher values of \(p\). The robustness and effectiveness of HI\textsuperscript{2} are further highlighted by its ability to maintain high performance even when other baselines show significant variability or lower results. Overall, the HI\textsuperscript{2} approach proves to be a highly reliable and efficient choice for haphazard inputs, making it a strong candidate for further applications and research.

We observe that the deep-learning models underperform compared to classical models in a8a, owing to a high-class imbalance (75.92\%). Still, the HI\textsuperscript{2} performs the best among the deep-learning models at each $p$ value across all metrics.

\begin{table}[t]
\caption{Results of all the ablation studies on magic04 dataset. }
\begin{center}
\begin{tabular}{lccc}
\toprule
 & \multicolumn{3}{c}{Balanced Accuracy} \\
 \cmidrule{2-4}
 & \(p\) = 0.25 & \(p\) = 0.5 & \(p\) = 0.75 \\
\midrule
\multicolumn{4}{c}{\underline{\textbf{Graphical Representation}}} \vspace{0.5em}\\
Pie Chart & 51.99$\pm$0.93  & 61.12$\pm$0.95 & 70.28$\pm$0.48 \\
Bar Graph (X marking) & \textbf{60.47$\pm$0.31} & \textbf{67.84$\pm$0.39} & \textbf{74.48$\pm$0.20}\\
HI\textsuperscript{2} (Only Bar Graph) & 60.04$\pm$0.26 & 67.13$\pm$0.37 & 74.34$\pm$0.09 \\

\midrule
\multicolumn{4}{c}{\underline{\textbf{Streaming Normalization}}} \vspace{0.5em} \\
Min-Max & 57.07$\pm$0.22 & 65.32$\pm$0.17 & 73.18$\pm$0.44 \\
HI\textsuperscript{2} (Z-score) & \textbf{60.04$\pm$0.26} & \textbf{67.13$\pm$0.37} & \textbf{74.34$\pm$0.09} \\

\midrule
\multicolumn{4}{c}{\underline{\textbf{Model-Agnostic}}} \vspace{0.5em} \\
ViT-Small &60.00$\pm$0.44 & 66.30$\pm$0.05 &72.46$\pm$0.65 \\
HI\textsuperscript{2} (ResNet-34) & \textbf{60.04$\pm$0.26} & \textbf{67.13$\pm$0.37} & \textbf{74.34$\pm$0.09} \\

\bottomrule
\end{tabular}
\label{tab:ablation}
\end{center}
\end{table}

\section{Ablation Study}
We perform ablation studies on the HI\textsuperscript{2} concept, considering balanced accuracy in accordance with previous literature \cite{HaphazardInputSurvey}. Here, we consider the magic04 dataset and adhere to the implementation details discussed in section \ref{sec:implementation_details} unless specified otherwise. The ablation studies are discussed next.

\subsection{Graphical Representation}

We choose bar graphs to represent the observed features of haphazard inputs. Here, we show the significance of our choice by presenting the problems of other graphical representations.

\paragraph{Pie Chart}
The pie chart is one of the simplest graphical representations that depict the relative proportion among features as areas of sectors. We represent the observed features by partitioning the 360 degrees of a pie based on the normalized values of the features. The colors serve as unique identities for features, similar to bar graphs. The comparison between HI\textsuperscript{2} on the pie chart and the bar graph is presented in the first block of Table \ref{tab:ablation}. The pie chart performs 5.46\%, 8.95\%, and 13.41\% poorer than the bar graph at $p = $ 0.75, 0.50, and 0.25, respectively. This poorer performance results from information loss due to inter-feature normalization to fit the data in 360 degrees. However, the pie chart visual representation still outperforms other baseline models (see Table \ref{tab:benchmarking}) by a significant margin on magic04 at \(p = 0.50\) and 0.75, showing the efficacy of visually representing the haphazard inputs and utilizing pretrained vision models.

\paragraph{Line Graphs} 
\label{sec:linegraphs}
The line graph is a simple and important data representation method, able to showcase the trend in features through time. This is evidenced in ViTST \cite{ViTST} in an offline learning setting. However, in online learning, the restriction of not storing data prohibits creating line graphs. Therefore, although a line graph can represent temporal information, it may not be employed for haphazard inputs.

\paragraph{Bar graph representing missing data, missing features, and obsolete features} The bar graph in HI\textsuperscript{2} represents only the observed features because it allows to represent a large number of features as discussed in section \ref{sec:ImageTransformation}. However, it may lose information by not depicting the unobserved features resulting from missing data, missing features, and obsolete features. Representing these unobserved features (by an ``X" marking in place of the bar) would require a large canvas size, followed by resizing them to 224\(\times\)224 dimensions. This resizing may lead to information loss. However, this representation has two benefits: (1) it preserves the position of the bar in addition to the unique feature color, thus aiding the vision models, and (2) it explicitly provides information about the unobserved features. The first block in Table \ref{tab:ablation} quantifies the benefit of the bar graph with X marking compared to the original HI\textsuperscript{2} on the magic04 dataset, where resizing is not required. However, if generalized for a large number of features (say 10k), the X-marking method will lose information. Therefore, in the trade-off between representing unobserved features or representing a large number of observed features, we advocate for the latter option. Both with and without an X marking approach gives superior results than the baseline, and provides a choice between them based on the use case.

\subsection{Model-Agnostic Property of HI\textsuperscript{2}} 
The model agnostic property of HI\textsuperscript{2} alleviates the need for specialized models in haphazard inputs, as demonstrated in Fig. \ref{fig:teaser}. This presents an opportunity to utilize pre-trained state-of-the-art existing deep learning models. This also allows a choice of the model based on the use case. For example, a large and complex model like ViT-Large or a simple and small model like ResNet-18 can be easily incorporated into the HI\textsuperscript{2} approach depending on the use cases.

To demonstrate this model-agnostic property of HI\textsuperscript{2}, in addition to ResNet-34, we also employ Vision Transformer (ViT) \cite{ViT} to handle haphazard inputs. We observed that ResNet-34 outperforms ViT-Small (Table \ref{tab:ablation}). However, ViT on the HI\textsuperscript{2} concept still outperforms all the baseline models on the magic04 dataset. This experiment reaffirms the benefit of using a CV-based model for haphazard inputs.

\subsection{Different Normalizations}

The Z-score normalization outperforms the Min-Max normalization by a decent margin (see Table \ref{tab:ablation}). This is because the Z-score can handle outliers. Moreover, the Z-score can adapt to the varying distribution nature of the streaming data by recalculating the mean and standard deviation as new data arrives. Still, Min-Max normalization outperforms all the baseline models, proving that irrespective of the normalization technique, the HI\textsuperscript{2} is superior to the baseline models.

\section{Time and Space Complexity}

The time required by HI\textsuperscript{2} culminates from the CV model and image transformation. We note that the time required by the CV model is always constant because of the fixed dimension of the image. Therefore, two datasets with different numbers of features and an equal number of instances will require the same amount of time by the CV model. The minuscule difference results from the image transformation. This is evident in SUSY and HIGGS (number of features is 8 and 21) datasets containing an equal number of instances, where HI\textsuperscript{2} requires almost the same amount of time (20566.35\(\pm\)205.84 and 21565.05\(\pm\)43.91) at $p = $ 0.25. Therefore, the time complexity of HI\textsuperscript{2} is not highly dependent on the number of features. The time required by HI\textsuperscript{2} on magic04 and a8a dataset at $p = $ 0.25 is 388.93\(\pm\)21.61 and 923.31\(\pm\)3.07, respectively. 

The space complexity of HI\textsuperscript{2} largely depends on the CV model and is constant. Thus, HI\textsuperscript{2} is scalable in terms of the number of features for both time and space complexity. This is an important property in the highly inconsistent input space.

\section{Conclusion}

In conclusion, HI\textsuperscript{2} is a model-agnostic approach that can handle haphazard inputs in online learning settings by transforming them into fixed-dimension images. HI\textsuperscript{2} leverages pre-trained vision models and demonstrates significant improvements over baselines across various datasets and metrics.

In our work, we also identify a few aspects of HI\textsuperscript{2} that can be worked upon. The bar graph representation of haphazard inputs does not preserve the temporal information as possible in line graphs. However, as discussed in section \ref{sec:linegraphs}, line graphs may not represent the haphazard inputs. Although HI\textsuperscript{2} performed better than other deep-learning models in the highly imbalanced a8a dataset, it underperformed compared to some of the classical methods. Therefore, future work could focus on preserving temporal information and better handling of class imbalance to enhance the efficacy of HI\textsuperscript{2}. We would also like to explore potential applications like space exploration, sub-cellular modeling, aircraft health monitoring, and precision agriculture using IoT that generates haphazard inputs.

\end{document}